%% file: acl_latex.tex
\documentclass[11pt]{article}

\usepackage[final]{acl}

\usepackage{times}
\usepackage{latexsym}
\usepackage[T1]{fontenc}
\usepackage[utf8]{inputenc}
\usepackage{microtype}
\usepackage{inconsolata}
\usepackage{booktabs}
\usepackage{amsmath,amssymb}
\usepackage{graphicx}
\usepackage{multirow}

\usepackage{tikz}
\usetikzlibrary{arrows.meta, positioning}
\colorlet{cPurple}{violet!70!black}
\usepackage{xcolor}
\usetikzlibrary{positioning,arrows.meta,fit,shapes.geometric,shadows,decorations.pathreplacing,calc}

\title{From Veracity to Diffusion: Adressing Operational Challenges in Moving From Fake-News Detection to Information Disorders}

\author{Francesco Paolo Savatteri, Chahan Vidal-Gorène, Florian Cafiero \\ Ecole nationale des chartes - PSL\\
         65 rue de Richelieu, 72002, Paris, France \\
         francesco.savatteri@chartes.psl.edu, chahan.vidalgorene@chartes.psl.eu florian.cafiero@chartes.psl.eu} 

\begin{document}
\maketitle

\begin{abstract}

A wide part of research on misinformation has relied lies on fake-news detection, a task framed as the prediction of veracity labels attached to articles or claims. Yet social-science research has repeatedly emphasized that information manipulation goes beyond fabricated content and often relies on amplification dynamics. This theoretical turn has consequences for operationalization in applied social science research. What changes empirically when prediction targets move from veracity to diffusion? And which performance level can be attained in limited resources setups ? In this paper we compare fake-news detection and virality prediction across two datasets, \textsc{EVONS} and \textsc{FakeNewsNet}. We adopt an evaluation-first perspective and examine how benchmark behavior changes when the prediction target shifts from veracity to diffusion. Our experiments show that fake-news detection is comparatively stable once strong textual embeddings are available, whereas virality prediction is much more sensitive to operational choices such as threshold definition and early observation windows. The paper proposes practical ways to operationalize lightweight, transparent pipelines for misinformation-related prediction tasks that can rival with state-of-the-art.
\end{abstract}

\input{schema.tex}

\section{Introduction}

Online misinformation has become a major concern across computational social science, media studies, and NLP \citep{Lazer2018, Srba2024credibility}. In computational work, this problem has most often been operationalized as \emph{fake-news detection}, that is, the prediction of veracity labels attached to articles, claims, posts, or rumors \citep{Zubiaga2018, Papageorgiou2024fakeNewsLLMs}.

At the same time, work in communication studies and policy has argued that problematic information ecosystems cannot be reduced to falsehood alone. Broader frameworks such as \emph{information disorder} emphasize that manipulation may involve misinformation, disinformation, malinformation, and strategic amplification, rather than fabricated content in a narrow sense \citep{WardleDerakhshan2017}.

This suggests that misinformation research should not focus exclusively on veracity-based targets. A complementary perspective is to model \emph{diffusion-oriented} outcomes such as virality, popularity, or spread \citep{Vosoughi2018, zhang2024viral_rumors_vulnerable_users}. These tasks are still often studied separately from fake-news detection. Yet combining the two perspectives is not merely a theoretical exercise: it reflects a practical necessity. Social media produces information at a scale that makes it really hard for journalists, citizens, or regulatory bodies to verify and debunk every identified piece of fake news (375 million posts in 24  hours only on Twitter; \citet{Pfeffer2023}). Any realistic counter-disinformation pipeline must therefore include a selection strategy that prioritizes the most consequential content to focus on. Virality prediction is a natural candidate for such a role: by estimating which items are most likely to spread, it can guide attention toward the subset of misinformation that poses the greatest risk of broad societal impact.

This paper addresses that question through a comparative empirical study on two resources, \textsc{EVONS} and \textsc{FakeNewsNet} \citep{krstovski2022evons, shu2020fakenewsnet}. We adopt an evaluation-first perspective and compare fake-news detection and virality prediction under a common experimental framework. Our goal is to examine how benchmark behavior changes when the target itself changes, and to design a robust enough lightweight pipeline that can be operational in a realistic framework for applied social sciences studies.

\section{Related Work}
\label{sec:related}

NLP research on misinformation has been organized around veracity-oriented tasks such as fake-news detection, rumor detection, and related classification problems \citep{Conroy2015, Shu2017, Zubiaga2018}. This literature has expanded methodologically toward multimodal fusion, graph neural networks, and LLM-based systems, but it still largely treats misinformation as a problem of predicting whether content is true or false \citep{Liu2020, alam2022survey, comito2023multimodal_survey, liu2023bridginggap, zeng2024learnfromlimited, phan2023gnn_fakenews_survey, chen2024llms_misinformation, Papageorgiou2024fakeNewsLLMs}. Closely related work has also moved toward source-level factuality and bias prediction, showing that misinformation research increasingly includes coarser-grained profiling tasks beyond article-level veracity classification \citep{Nakov2024factualityBias}.Social science research however argues that the notion of ``fake news'' is itself too narrow. Broader perspectives emphasize information disorder, disinformation ecosystems, media manipulation, and amplification mechanisms \citep{WardleDerakhshan2017, HLEG2018, BennettLivingston2018, MarwickLewis2017}. Recent survey argues for moving toward richer credibility-oriented formulations \citep{Srba2024credibility}.

A second line of work focuses on diffusion, virality, and propagation dynamics. Studies of online news circulation show that false and true information differ not only in content but also in how they spread \citep{Vosoughi2018}. Resources such as \textsc{EVONS} and \textsc{FakeNewsNet} make it possible to study veracity and diffusion within related empirical settings, while recent work also models viral rumor spread and vulnerable users as explicit predictive targets \citep{zhang2024viral_rumors_vulnerable_users}. Studies on virality prediction on \textsc{FakeNewsNet} splits rarely take into account the sequential structure of the data \citep{EstebanBravo2024}, which allows for a more accurate simulation of the spread of an online narrative, rather than a single publication. Direct comparisons between veracity prediction and diffusion prediction remain uncommon.  

Benchmark results depend strongly on task definition, dataset construction, and measurement choices \citep{Chang2024}. This is especially important in this field, as datasets differ in label provenance, modality, and access to platform-derived signals, and as API restrictions increasingly affect transparency and reproducibility \citep{Ferrara2025}. Recent work also shows that misinformation detection is highly sensitive to data quality and evaluation design, and argues that standard metrics can be misleading when dataset quality is weak or poorly documented \citep{Thibault2025guide}. 

\section{Tasks and Data}
\label{sec:tasks-data}

Our goal is to compare two families of prediction tasks: \emph{veracity prediction} and \emph{diffusion prediction}. In fake-news detection, the objective is to predict the dataset-provided veracity label associated with each instance. In virality prediction, the objective is to predict whether the engagement associated with an instance exceeds a dataset-specific threshold. The first task therefore relies on externally provided labels, whereas the second depends on an explicit operationalization of virality.

\subsection{Datasets}

\subsubsection{\textsc{EVONS}}

\textsc{EVONS} was introduced by \citet{krstovski2022evons} to support the joint study of fake news and virality. It contains news articles with veracity labels and article-level engagement statistics. For fake-news detection, we use the dataset's binary veracity labels. For virality prediction, we define a binary label using a high-quantile threshold over the engagement distribution, following the logic of rare-event detection.

\subsubsection{\textsc{FakeNewsNet}}

\textsc{FakeNewsNet} combines news content with social-context and temporal information collected from Twitter \citep{shu2020fakenewsnet}. It includes article-level veracity labels together with propagation-related information, making it suitable for both experiments.

\section{Methods}
\label{sec:methods}

\subsection{Text Representations}
\label{subsec:text-representations}

All experiments are based on fixed dense embeddings computed from the textual content of each instance. For \textsc{EVONS}, we embed the article title and description (or caption) separately and concatenate the resulting vectors into a single article-level representation. For \textsc{FakeNewsNet}, each instance is represented by a mean-pooled fixed-size text embedding computed from the textual content associated with that news item in the experimental representation. The downstream models therefore consume one embedded vector per instance.

We compare two embedding backbones:

\begin{itemize}
    \item a RoBERTa-based representation, denoted \texttt{bert} in the exported result files, with dimensionality 768;
    \item a Mistral-based representation, denoted \texttt{mistral}, with dimensionality 1024.
\end{itemize}

The purpose of this comparison is not to benchmark foundation models \textit{per se}, but to test how our conclusions depend on the choice of embedding space.

\subsection{Compared Model Families}
\label{subsec:model-families}

Our main experiments compare lightweight supervised classifiers operating on the fixed text embeddings described above. The compared model families are multilayer perceptrons (MLP), logistic regression, random forests, and gradient-boosted trees (XGBoost). The exported result files identify each system by combining the classifier family and the embedding backbone (e.g., \texttt{mlp\_bert}, \texttt{rf\_mistral}, \texttt{xgboost\_bert}).

For \textsc{EVONS} virality, we compare lightweight ways of incorporating source-related information. In addition to a text-only MLP (\texttt{mlp\_text}), we evaluate three source-aware variants: \texttt{mlp\_source}, which encodes sources as categorical variables; \texttt{mlp\_avg\_eng}, which incorporates the average engagement of each source as an additional feature; and gated fusion models combining text embeddings with engagement information (\texttt{gating\_bert}, \texttt{gating\_mistral}). For \texttt{mlp\_avg\_eng} and the gating fusion models, the intuition is that articles from outlets with average higher engagement are more likely to go viral; average engagement values are computed per fold exclusively on training set data to prevent data leakage.

\subsection{Virality as an Operationalized Target}
\label{subsec:virality-operationalization}

A central methodological point of the paper is that virality is not a naturally binary property, but the result of thresholding a continuous engagement distribution. Let $\{e_i\}_{i=1}^{N}$ denote the engagement values in a dataset. For a quantile $q$, the corresponding threshold is
\[
\tau_q = Q_q(\{e_i\}),
\]
and the resulting virality label is
\[
y_i^{(\mathrm{vir})} = \mathbf{1}[e_i \ge \tau_q].
\]

Different values of $q$ therefore define different prediction problems. A median split yields a balanced classification task, whereas high-quantile thresholds produce rare-event detection settings with strong class imbalance. To examine this, we run a virality-sensitivity analysis on \textsc{FakeNewsNet}. Using ordered propagation data, we evaluate thresholds at $
q \in \{0.50, 0.75, 0.90, 0.95\}$.
In this audit, engagement is computed from tweet-level \texttt{favorite\_count} values. For each propagation sequence, we compute both total engagement
\[
E_i = \sum_{j=1}^{L_i} \ell_{ij}
\]
and prefix-based engagement
\[
E_i^{(k)} = \sum_{j=1}^{k} \ell_{ij},
\]
where $\ell_{ij}$ is the engagement associated with post $j$ in sequence $i$ and $k$ is the number of early posts observed. We evaluate prefix lengths $ k \in \{1,3,5,10\} $ which allows us to study how much information about final virality is already available from the early part of the diffusion sequence.

\subsection{Evaluation Protocol}
\label{subsec:evaluation-protocol}

The main experiments are evaluated by stratified 10-fold cross-validation.

Across all settings, we report the same standard binary-classification metrics:Accuracy, F1, Precision, Recall, ROC-AUC.
Because several virality settings are class-imbalanced, we do not interpret accuracy in isolation. F1 is used as the main summary metric, while ROC-AUC captures ranking quality independently of a fixed decision threshold.

For the auxiliary virality audits, we additionally report the threshold value, class prevalence at each quantile, and prefix-based AUC.

\subsection{Paired Statistical Comparison}
\label{subsec:statistical-comparison}

To avoid overinterpreting small differences between models, we complement average scores with fold-level paired statistical comparisons. For each dataset--task--metric combination, models are compared on matched test folds. Let $s_f(A)$ and $s_f(B)$ denote the score of models $A$ and $B$ on fold $f$, and define the fold-wise difference as
\[
d_f = s_f(A) - s_f(B).
\]

For each pair of models, we report the mean difference across matched folds; bootstrap confidence interval obtained by resampling the fold-wise differences; Cliff's delta as a non-parametric effect-size estimate; a sign-flip $p$-value; Holm-adjusted $p$-values for multiple comparisons within each dataset--task--metric family.

\section{Results}
\label{sec:results}

Unless otherwise specified, scores correspond to means across the 10 folds of stratified cross-validation.

\subsection{Fake-News Detection}

Table~\ref{tab:fakenews-results} reports fake-news detection results on \textsc{EVONS} and \textsc{FakeNewsNet}. Across both datasets, performance is strong and relatively stable once high-quality text embeddings are available.

On \textsc{EVONS}, all models achieve high scores. The best configuration is the MLP using Mistral embeddings (F1 = 0.988; ROC-AUC = 0.999), followed by the BERT-based MLP (F1 = 0.981). Linear and tree-based baselines remain competitive but weaker than the best neural configuration.

On \textsc{FakeNewsNet}, the best model is a random forest over BERT embeddings (F1 = 0.906; ROC-AUC = 0.970). The two MLP variants also perform strongly (F1 = 0.903 with Mistral and 0.895 with BERT). More generally, the strongest models remain within a relatively narrow performance range, which suggests that fake-news detection is comparatively stable across classifier families once strong textual representations are available.

\begin{table*}[t]
\centering
\small
\caption{Fake-news detection results (means across 10 cross-validation folds).}
\label{tab:fakenews-results}
\begin{tabular}{llccccc}
\toprule
Dataset & Model & Accuracy & F1 & Precision & Recall & ROC-AUC \\
\midrule
\multirow{8}{*}{EVONS}
& mlp\_mistral      & \textbf{0.989} & \textbf{0.988} & \textbf{0.988} & \textbf{0.989} & \textbf{0.999} \\
& mlp\_bert         & 0.982 & 0.981 & 0.984 & 0.978 & 0.999 \\
& logreg\_mistral   & 0.975 & 0.973 & 0.978 & 0.967 & 0.997 \\
& xgboost\_bert     & 0.972 & 0.970 & 0.975 & 0.964 & 0.997 \\
& logreg\_bert      & 0.971 & 0.968 & 0.974 & 0.963 & 0.996 \\
& xgboost\_mistral  & 0.956 & 0.955 & 0.957 & 0.953 & 0.994 \\
& rf\_bert          & 0.949 & 0.944 & 0.958 & 0.930 & 0.992 \\
& rf\_mistral       & 0.922 & 0.922 & 0.919 & 0.926 & 0.981 \\
\midrule
\multirow{8}{*}{FakeNewsNet}
& rf\_bert          & \textbf{0.946} & \textbf{0.906} & 0.942 & \textbf{0.874} & 0.970 \\
& mlp\_mistral      & 0.945 & 0.903 & \textbf{0.959} & 0.857 & \textbf{0.981} \\
& mlp\_bert         & 0.940 & 0.895 & 0.941 & 0.857 & 0.975 \\
& xgboost\_bert     & 0.930 & 0.880 & 0.906 & 0.860 & 0.975 \\
& rf\_mistral       & 0.929 & 0.872 & 0.955 & 0.806 & 0.956 \\
& logreg\_bert      & 0.923 & 0.865 & 0.880 & 0.857 & 0.975 \\
& xgboost\_mistral  & 0.914 & 0.859 & 0.887 & 0.834 & 0.972 \\
& logreg\_mistral   & 0.844 & 0.721 & 0.967 & 0.583 & 0.951 \\
\bottomrule
\end{tabular}
\end{table*}

\subsection{Virality Prediction}

Virality prediction yields a markedly different empirical picture (Table~\ref{tab:virality-results}).

On \textsc{EVONS}, performance is low overall and highly uneven across models. The best configuration is the Mistral-based gated model (F1 = 0.312; ROC-AUC = 0.881). The two source-aware variants perform substantially worse in F1 despite reasonably strong AUC values (F1 = 0.087 for \texttt{mlp\_avg\_eng} and F1 = 0.084 for \texttt{mlp\_source}). The BERT-based gating model and the text-only MLP collapse almost completely at the classification threshold, with F1 values close to zero despite AUC values of 0.868 and 0.828 respectively. This indicates that, under the virality definition used in \textsc{EVONS}, ranking ability and thresholded classification performance diverge sharply.

On \textsc{FakeNewsNet}, the situation is quite different. After aggregating fold-level results by model, all systems fall into a narrow band, with F1 values ranging from 0.740 to 0.777. The best model is XGBoost with BERT embeddings (F1 = 0.777; ROC-AUC = 0.861), followed closely by \texttt{mlp\_bert} and \texttt{logreg\_mistral} (both around F1 = 0.772). Unlike in \textsc{EVONS}, no single model clearly dominates, and virality prediction behaves more like a balanced classification problem.

\begin{table*}[t]
\centering
\small
\caption{Virality prediction results. For \textsc{FakeNewsNet}, scores are aggregated from the fold-level merged results.}
\label{tab:virality-results}
\begin{tabular}{llccccc}
\toprule
Dataset & Model & Accuracy & F1 & Precision & Recall & ROC-AUC \\
\midrule
\multirow{5}{*}{EVONS}
& gating\_mistral & 0.950 & \textbf{0.312} & 0.525 & \textbf{0.229} & \textbf{0.881} \\
& mlp\_avg\_eng   & \textbf{0.951} & 0.087 & 0.631 & 0.048 & 0.868 \\
& mlp\_source     & 0.951 & 0.084 & \textbf{0.667} & 0.047 & 0.870 \\
& gating\_bert    & 0.950 & 0.006 & 0.063 & 0.003 & 0.868 \\
& mlp\_text       & 0.950 & 0.006 & 0.313 & 0.003 & 0.828 \\
\midrule
\multirow{8}{*}{FakeNewsNet}
& xgboost\_bert    & \textbf{0.770} & \textbf{0.777} & 0.754 & 0.803 & 0.861 \\
& mlp\_bert        & 0.748 & 0.772 & 0.704 & \textbf{0.859} & 0.806 \\
& logreg\_mistral  & 0.755 & 0.772 & 0.723 & 0.828 & 0.808 \\
& logreg\_bert     & 0.749 & 0.766 & 0.720 & 0.819 & 0.809 \\
& rf\_bert         & 0.761 & 0.765 & \textbf{0.755} & 0.776 & \textbf{0.862} \\
& mlp\_mistral     & 0.750 & 0.760 & 0.735 & 0.797 & 0.809 \\
& xgboost\_mistral & 0.736 & 0.740 & 0.731 & 0.750 & 0.831 \\
& rf\_mistral      & 0.749 & 0.740 & 0.768 & 0.716 & 0.849 \\
\bottomrule
\end{tabular}
\end{table*}

\subsection{Virality Threshold Sensitivity}

To complement the main model comparisons, we run an auxiliary virality audit on \textsc{FakeNewsNet}. In this audit, instances are split into the subsets labeled as \texttt{real} and \texttt{fake} by the veracity annotations, and virality thresholds are computed separately within each subset. Table~\ref{tab:thresholds} shows how the engagement threshold changes as the virality quantile increases in these two subsets.

In the \texttt{real} subset, the threshold rises from 19.5 likes at the median to 59{,}315.15 likes at the 95th percentile, while the positive class shrinks from 50.0\% to 5.2\% of the data. In the \texttt{fake} subset, the threshold rises from 46.5 to 2{,}828.2 likes, with the same reduction in prevalence. Changing the threshold therefore does not merely rebalance the data slightly: it changes the substantive meaning of what counts as ``viral.''

\begin{table}[t]
\centering
\small
\caption{Virality thresholds in the auxiliary \textsc{FakeNewsNet} audit files.}
\label{tab:thresholds}
\begin{tabular}{lccc}
\toprule
Subset & Quantile & Threshold & Positive rate \\
\midrule
Real & 0.50 & 19.5 & 0.50 \\
Real & 0.75 & 1514.8 & 0.25 \\
Real & 0.90 & 14371.0 & 0.10 \\
Real & 0.95 & 59315.2 & 0.05 \\
\midrule
Fake & 0.50 & 46.5 & 0.50 \\
Fake & 0.75 & 295.3 & 0.25 \\
Fake & 0.90 & 1553.9 & 0.10 \\
Fake & 0.95 & 2828.2 & 0.05 \\
\bottomrule
\end{tabular}
\end{table}

\subsection{Early-Signal Predictability}

\begin{figure*}[t]
    \centering
    \begin{minipage}[t]{0.49\textwidth}
        \centering
        \includegraphics[width=\linewidth]{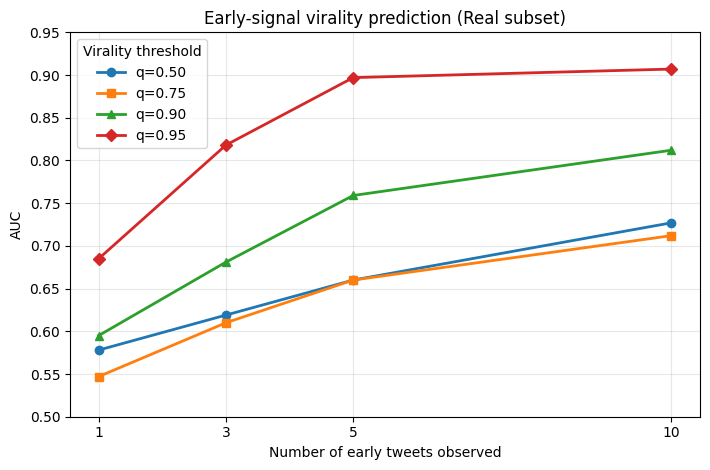}
    \end{minipage}\hfill
    \begin{minipage}[t]{0.49\textwidth}
        \centering
        \includegraphics[width=\linewidth]{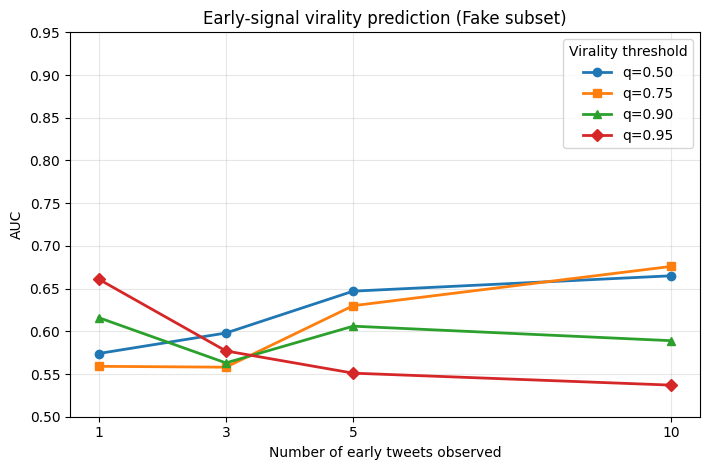}
    \end{minipage}
    \caption{Early-signal virality prediction from prefix engagement only. Left: \texttt{real} subset.  Right: \texttt{fake} subset.}
    \label{fig:early-signal}
\end{figure*}

Using the same auxiliary audit split, we evaluate how much information about final virality is already contained in the early part of the propagation sequence. Figure~\ref{fig:early-signal} reports AUC values obtained from prefix engagement only.

In the \texttt{real} subset, early signals become substantially more informative as more tweets are observed. For the median split, AUC rises from 0.578 with the first tweet to 0.727 with the first 10 tweets; for the 95th-percentile label, it rises from 0.685 to 0.907. In the \texttt{fake} subset, the same pattern is weaker and less stable. For example, the median-based label rises from 0.574 to 0.665, but the 95th-percentile label declines from 0.661 at one tweet to 0.537 at 10 tweets. Early engagement therefore carries meaningful information, but its predictive value depends strongly on the subset and on the virality threshold.

%\begin{table}[t]
%\centering
%\small
%\caption{Early-signal virality prediction (AUC) from prefix engagement only.}
%\label{tab:early}
%\resizebox{\columnwidth}{!}{%
%\begin{tabular}{lccccc}
%\toprule
%Subset & Quantile & AUC@1 & AUC@3 & AUC@5 & AUC@10 \\
%\midrule
%Real & 0.50 & 0.578 & 0.619 & 0.660 & 0.727 \\
%Real & 0.75 & 0.547 & 0.610 & 0.660 & 0.712 \\
%Real & 0.90 & 0.595 & 0.681 & 0.759 & 0.812 \\
%Real & 0.95 & 0.685 & 0.818 & 0.897 & 0.907 \\
%\midrule
%Fake & 0.50 & 0.574 & 0.598 & 0.647 & 0.665 \\
%Fake & 0.75 & 0.559 & 0.558 & 0.630 & 0.676 \\
%Fake & 0.90 & 0.616 & 0.563 & 0.606 & 0.589 \\
%Fake & 0.95 & 0.661 & 0.577 & 0.551 & 0.537 \\
%\bottomrule
%\end{tabular}
%}
%\end{table}

\subsection{Fold-level Statistical Comparison}
\label{subsec:results-stats}

We conducted paired fold-level statistical comparisons on F1. For each dataset--task setting, we compare the best model by mean F1 with its closest competitor, defined here as the second-best model by mean F1. Table~\ref{tab:stats-runnerup} reports the mean fold-wise F1 difference, the corresponding bootstrap confidence interval, and the Holm-corrected $p$-value. For fake-news detection, the differences between the strongest models remain small and do not survive Holm correction, both on \textsc{EVONS} and on \textsc{FakeNewsNet}. Likewise, for \textsc{FakeNewsNet} virality, the best model is only marginally ahead of its closest competitor and the difference is not statistically significant. By contrast, the \textsc{EVONS} virality setting shows a large and statistically robust separation: \texttt{gating\_mistral} significantly outperforms the second-best model after Holm correction. 

\begin{table*}[t]
\centering
\small
\caption{Paired fold-level F1 comparisons between the best model and the runner-up in each dataset/task setting. $\Delta$F1 is the mean fold-wise difference (best minus runner-up).}
\label{tab:stats-runnerup}
\begin{tabular}{llcccccc}
\toprule
Dataset & Task & Best model & Runner-up & $\Delta$F1 & 95\% CI & Holm-$p$ & Sig. \\
\midrule
EVONS & Fake-news & \texttt{mlp\_mistral} & \texttt{mlp\_bert} & 0.0078 & [0.0065, 0.0091] & 0.0547 & no \\
EVONS & Virality & \texttt{gating\_mistral} & \texttt{mlp\_avg\_eng} & 0.2253 & [0.1701, 0.2753] & 0.0195 & yes \\
FakeNewsNet & Fake-news & \texttt{rf\_bert} & \texttt{mlp\_mistral} & 0.0036 & [-0.0182, 0.0266] & 1.0000 & no \\
FakeNewsNet & Virality & \texttt{xgboost\_bert} & \texttt{mlp\_bert} & 0.0052 & [-0.0142, 0.0266] & 1.0000 & no \\
\bottomrule
\end{tabular}
\end{table*}

\section{Discussion}
\label{sec:discussion}

Fake-news detection behaves as a comparatively stable task in our setting. Across both \textsc{EVONS} and \textsc{FakeNewsNet}, once strong textual embeddings are available, several downstream models reach high performance and remain relatively close to one another. On \textsc{EVONS}, performance is near ceiling for the strongest systems; on \textsc{FakeNewsNet}, the best models again cluster within a narrow range. This does not imply that model choice is irrelevant, but it suggests that substantial part of the discriminative signal is already captured by the text representation. In that sense, fake-news detection appears to be a well-behaved benchmark: labels are externally provided, the prediction target is fixed, and the strongest systems yield convergent results.

The diffusion-oriented setting is more revealing. Virality prediction is not simply a harder version of fake-news detection; it is a more contingent task whose empirical behavior depends strongly on operationalization. This is visible first in the contrast between datasets. On \textsc{EVONS}, virality behaves like a rare-event problem with sharp instability across models: the best system remains modest in F1, several competitors collapse at the default classification threshold, and ROC-AUC diverges markedly from thresholded performance. On \textsc{FakeNewsNet}, by contrast, virality prediction under the median-based construction is much more regular, with all models falling into a narrow performance band. The important point is therefore not that one dataset is ``easy'' and the other ``hard,'' but that virality is not a single natural target. Different operational definitions induce substantively different prediction problems.

The auxiliary analyses reinforce this interpretation. Changing the virality quantile does not simply rebalance the classes; it changes what counts as viral in substantive terms. In the \texttt{real} subset of \textsc{FakeNewsNet}, the threshold moves from 19.5 likes at the median to more than 59{,}000 at the 95th percentile; in the \texttt{fake} subset, it moves from 46.5 to more than 2{,}800. These are not minor calibration choices. They alter class prevalence, decision difficulty, and the meaning of the positive class itself. From this perspective, benchmark results in diffusion-oriented prediction partly measure model quality, but they also reflect target construction. This is the central claim of the paper: once prediction shifts from veracity to diffusion, evaluation becomes inseparable from operationalization.

The early-signal audit points in the same direction. Early engagement can indeed be informative, sometimes strongly so, but its usefulness is uneven across subsets and thresholds. In the \texttt{real} subset, predictive signal increases clearly as more early posts are observed, especially for the most extreme virality labels. In the \texttt{fake} subset, however, this pattern weakens and can even reverse at the highest quantile. This matters for two reasons. First, it shows that observation window is itself part of task design. Second, it cautions against treating ``early virality prediction'' as a uniform capability. What can be predicted early depends on what final outcome is being predicted and on which subset of the data is considered.

A second contribution of the paper is practical. The experiments show that lightweight pipelines based on fixed embeddings and standard downstream classifiers are sufficient both to obtain strong fake-news detection results and to surface meaningful contrasts in diffusion-oriented settings. This is useful methodologically because it allows us to study task formulation without entangling the analysis with highly complex architectures. It is also useful empirically: although our objective is not to claim a new state of the art, our best fake-news result on \textsc{FakeNewsNet} exceeds the F1 reported by \citet{jiang2024model}, even if direct comparison should remain cautious because splits, preprocessing, and available inputs may differ. At minimum, this suggests that transparent and resource-efficient pipelines remain competitive baselines for misinformation-related prediction.

Moving beyond fake-news detection is necessary and, from an applied standpoint, urgent. The volume of content circulating on social media makes exhaustive verification unfeasible for any actor engaged in counter-disinformation work. Ideally, triage would be guided by estimated persuasive impact: prioritizing the items most likely to shift beliefs or reinforce misconceptions in the audience. Yet opinion change is notoriously difficult to measure at scale and in near-real time, making it an impractical operational target. Virality serves in this sense as a tractable proxy: not a perfect one, but an accessible signal that can inform prioritization within an automated pipeline. Our results, though, show that this move should not be treated as a simple extension of the same benchmark logic. Diffusion-oriented prediction requires more explicit reporting of threshold choice, class prevalence, and observation regime. Without that, evaluation risks conflating predictive performance with design decisions about what exactly is being predicted.

\section{Limitations}
The study remains limited in scope. It relies on two datasets only, and virality is operationalized through engagement thresholds rather than richer cascade- or network-based definitions. The paper therefore does not claim that virality is a sufficient proxy for manipulation, nor that diffusion-oriented tasks should replace veracity-based ones. Its contribution is narrower, but we think important: it shows that benchmark behavior changes qualitatively when the target moves from veracity to diffusion, and that this change is itself an object of methodological analysis. A natural next step would be to compare several operationalizations of diffusion within the same dataset and to extend the framework toward richer definitions of amplification, including repost structure, cascade shape, or coordinated activity.

\section*{Reproducibility and sharing.}
We release the configuration files and feature-extraction code necessary to reproduce the reported results, subject to dataset licenses and platform policies. For resources that cannot be redistributed directly (e.g., hydrated tweets), we document the reconstruction procedure and the relevant API endpoints so that authorized researchers can rebuild the inputs independently. Code, configuration files, and documentation are available on our \href{https://github.com/article-disinfo/css-acl-disinfo-virality}{GitHub repository}.

\bibliography{custom}
\end{document}

%% file: schema.tex
\definecolor{cBlue}{HTML}{2176AE}
\definecolor{cOrange}{HTML}{E07B39}
\definecolor{cGreen}{HTML}{2D9B6E}
\definecolor{cRed}{HTML}{C0392B}
\definecolor{cGray}{HTML}{5C6773}
\definecolor{cLightBlue}{HTML}{D9EAF7}
\definecolor{cLightOrange}{HTML}{FDEBD7}
\definecolor{cLightGray}{HTML}{F0F1F3}
\colorlet{cPurple}{violet!65!black}

\begin{figure}[th!]
\centering
\resizebox{\columnwidth}{!}{%
\begin{tikzpicture}[
  font=\sffamily\small,
  >=Stealth,
  enc/.style={rounded corners=4pt, draw=#1!70, fill=#1!8, text=#1!80!black,
    align=center, minimum width=28mm, minimum height=11mm,
    line width=1pt, font=\sffamily\small},
  reprbar/.style={rounded corners=2pt, draw=#1!60, fill=#1!18,
    minimum height=5mm, line width=0.8pt},
  clf/.style={rounded corners=4pt, draw=#1!65, fill=#1!8, text=#1!80!black,
    align=center, minimum width=32mm, minimum height=10mm,
    line width=0.9pt, font=\sffamily\scriptsize},
  task/.style={rounded corners=5pt, draw=#1!70, fill=#1!10, text=#1!80!black,
    align=center, minimum width=32mm, minimum height=13mm,
    line width=1.3pt, font=\sffamily\bfseries\small},
  dset/.style={rounded corners=3pt, draw=cGray!55, fill=cLightGray,
    align=center, minimum width=26mm, minimum height=7mm,
    line width=0.7pt, font=\sffamily\scriptsize},
  result/.style={rounded corners=6pt, draw=#1!65, fill=#1!10, text=#1!80!black,
    align=center, minimum width=32mm, minimum height=28mm, line width=1.4pt},
  arr/.style={->, line width=1.1pt, color=#1},
  darr/.style={->, dashed, line width=0.75pt, color=cGray!70},
  parr/.style={->, dashed, line width=0.9pt, color=cPurple},
  lbl/.style={font=\sffamily\tiny\bfseries, text=cGray!70},
]

%% ═══ R0 — INPUT (centred, full width) ════════════════════════════════════════
\node[rounded corners=6pt, draw=cGray!55, fill=cLightGray,
      align=center, minimum width=56mm, minimum height=14mm, line width=1pt]
  (input) at (0, 0)
  {\textbf{News article}\\
   \scriptsize title · body · captions\enspace+\enspace engagement stats};
\node[lbl, left=3mm of input] {INPUT};

%% ═══ R1 — ENCODERS (x=±2.8, y=−2.2) ════════════════════════════════════════
\node[enc=cBlue]   (encR) at (-2.8, -2.2) {RoBERTa\\\scriptsize encoder};
\node[enc=cOrange] (encM) at ( 2.8, -2.2) {Mistral\\\scriptsize encoder};
\draw[arr=cBlue!70]   (input.south) -- ++(0,-0.25) -| (encR.north);
\draw[arr=cOrange!70] (input.south) -- ++(0,-0.25) -| (encM.north);
\node[lbl] at (0, -1.35) {ENCODE};

%% ═══ R2 — EMBEDDING VECTORS (visual mini-vectors, y=−4.0) ════════════════════
%% RoBERTa d=768 → draw 12 mini-bars    Mistral d=1024 → draw 16 mini-bars
\coordinate (vRstart) at (-3.8, -4.0);
\foreach \i in {0,...,11} {
  \fill[cBlue!40, rounded corners=0.5pt]
    ($(vRstart)+(\i*0.12,0)$) rectangle ++(0.08,0.18);
}
\node[font=\ttfamily\tiny, text=cBlue!70, right] at ($(vRstart)+(1.5,0.09)$) {$d{=}768$};
\node[font=\ttfamily\tiny, text=cBlue!60, above=2pt] at ($(vRstart)+(0.5,0.18)$) {\texttt{[CLS]}};

\coordinate (vMstart) at (1.0, -4.0);
\foreach \i in {0,...,15} {
  \fill[cOrange!40, rounded corners=0.5pt]
    ($(vMstart)+(\i*0.12,0)$) rectangle ++(0.08,0.18);
}
\node[font=\ttfamily\tiny, text=cOrange!70, right] at ($(vMstart)+(2.0,0.09)$) {$d{=}1024$};
\node[font=\ttfamily\tiny, text=cOrange!60, above=2pt] at ($(vMstart)+(1.0,0.18)$) {\texttt{mean-pool}};

%% Anchor nodes for arrows (invisible, centred on mini-vector groups)
\coordinate (vR) at ($(vRstart)+(0.7, 0.09)$);
\coordinate (vM) at ($(vMstart)+(0.95, 0.09)$);

\draw[arr=cBlue!70]   (encR.south) -- ([yshift=0.18cm]encR.south |- vR);
\draw[arr=cOrange!70] (encM.south) -- ([yshift=0.18cm]encM.south |- vM);
\node[lbl] at (0, -3.3) {REPR.};

%% ═══ R3 — CLASSIFIERS: one box per lane (x=±2.8, y=−6.0) ════════════════════
%% Single box per lane → no same-row overlap possible
\node[clf=cBlue]   (clfV) at (-2.8, -6.0) {MLP · LR · RF · XGBoost};
\node[clf=cOrange] (clfD) at ( 2.8, -6.0) {MLP · LR · RF · XGBoost};
\draw[arr=cBlue!70]   ([yshift=-0.18cm]encR.south |- vR.south) -- (clfV.north);
\draw[arr=cOrange!70] ( [yshift=-0.18cm]encM.south |- vM.south) -- (clfD.north);
\node[lbl] at (0, -5.3) {CLASSIF.};

%% Gating node: centred, y=−7.6 (clearly below classifiers bottom ≈−6.5) ──────
\node[clf=cPurple] (clfG) at (0, -7.6)
  {\textcolor{cPurple!80!black}{gating fusion}\enspace\tiny eng.\,stats $\oplus$ text};
%% purple bypass: vM east → far right → down to clfG
\draw[parr] ([xshift=-1cm]vM.west) -- ++(-0.3,0) coordinate (tmp) -- (tmp |- clfG.north);

%% ═══ R4 — TASKS (x=±2.8, y=−9.2) ══════════════════════════════════════════
\node[task=cBlue]   (tA) at (-2.8, -9.2)
  {Veracity\\Prediction\\[1pt]\scriptsize labels from dataset};
\node[task=cOrange] (tB) at ( 2.8, -9.2)
  {Diffusion /\\Virality Pred.\\[1pt]\scriptsize threshold $\tau_q$ built};
\draw[arr=cBlue!80]   (clfV.south) -- (tA.north);
\draw[arr=cOrange!80] (clfD.south) -- (tB.north);
\draw[parr]           (tmp |- clfG.south) -- ++(0,-1.1) -- (tB.west);
\node[lbl] at (0, -8.3) {TASK};

%% ═══ R5 — DATASETS (absolute x, y=−11.2) ════════════════════════════════════
%% Lane A: x=−3.9 and −1.7   Lane B: x=+1.7 and +3.9
%% Node width=26mm (13mm half) — gap between A nodes: 2.2cm−1.3cm−1.3cm=−0.4→ need more
%% Use x=−4.2 and −1.4 (gap=2.8cm, node=2.6cm → 0.2cm clearance each side ✓)
\node[dset] (evA) at (-4.0, -11.2) {\textsc{EVONS}};
\node[dset] (fnA) at (-1.6, -11.2) {\textsc{FakeNewsNet}};
\node[dset] (evB) at ( 1.6, -11.2) {\textsc{EVONS}};
\node[dset] (fnB) at ( 4.0, -11.2)
  {\textsc{FakeNewsNet}\\[-2pt]\tiny$\tau_q{\in}\{.50,.75,.90,.95\}$};
\draw[darr] (tA.south) -- ++(0,-0.35) -| (evA.north);
\draw[darr] (tA.south) -- ++(0,-0.35) -| (fnA.north);
\draw[darr] (tB.south) -- ++(0,-0.35) -| (evB.north);
\draw[darr] (tB.south) -- ++(0,-0.35) -| (fnB.north);

%% ═══ R6 — FINDINGS (x=±2.8, y=−13.8) ══════════════════════════════════════
\node[result=cGreen] (findA) at (-2.8, -13.8)
  {\textcolor{cGreen!70!black}{\bfseries\normalsize Stable}\\[4pt]
   \scriptsize F1\,=\,0.90--0.99\\[1pt]
   \scriptsize $\Delta$F1\,$\approx$\,0.008 (n.s.)\\[1pt]
   \scriptsize AUC\,$\approx$\,F1\\[4pt]
   \scriptsize\itshape text repr.\ dominates};
\node[result=cRed] (findB) at ( 2.8, -13.8)
  {\textcolor{cRed!80!black}{\bfseries\normalsize Sensitive}\\[4pt]
   \scriptsize F1\,=\,0.006--0.78\\[1pt]
   \scriptsize AUC\,$\gg$\,F1 (diverge)\\[1pt]
   \scriptsize $\tau_q$ drives outcome\\[4pt]
   \scriptsize\itshape target design\,=\,results};
\draw[arr=cGreen] (evA.south) -- ++(0,-0.35) -| (findA.north);
\draw[arr=cGreen] (fnA.south) -- ++(0,-0.35) -| (findA.north);
\draw[arr=cRed]   (evB.south) -- ++(0,-0.35) -| (findB.north);
\draw[arr=cRed]   (fnB.south) -- ++(0,-0.35) -| (findB.north);
\node[lbl] at (0, -12.6) {FINDING};

\end{tikzpicture}
}% end resizebox
\caption{\textbf{Technical pipeline overview.}
Both tasks share the same backbone (RoBERTa $d{=}768$ or Mistral $d{=}1024$)
and classifiers (MLP, LR, RF, XGBoost).
For \textsc{EVONS} virality, engagement stats enter via a gating pathway (purple).
Veracity prediction yields stable results; virality is highly sensitive to $\tau_q$.}
\label{fig:teaser}
\end{figure}